\documentclass[preprint]{article}
\usepackage[table,xcdraw]{xcolor}
\usepackage{colortbl}
\usepackage{microtype}
\usepackage{graphicx}
\usepackage{subfigure}
\usepackage{booktabs}
\usepackage{cite}
\usepackage{amsmath,amssymb,amsfonts}
\usepackage{algorithmic}
\usepackage{textcomp}
\usepackage{xcolor}
\usepackage{makecell}
\usepackage{multirow}
\usepackage{diagbox}
\usepackage{balance}
\usepackage[flushleft]{threeparttable}
\usepackage[utf8]{inputenc} 
\usepackage[T1]{fontenc}    
\usepackage{hyperref}       
\usepackage{url}            
\usepackage{booktabs}      
\usepackage{amsfonts}       
\usepackage{nicefrac}      
\usepackage{microtype}
\usepackage{multirow}
\usepackage{listings}
\usepackage{adjustbox}
\usepackage{diagbox}
\usepackage{algorithm}
\usepackage{booktabs}
\usepackage{enumitem}
\usepackage{amsmath,amssymb}

\usepackage{pifont}

\usepackage{tcolorbox}

\newtcolorbox{promptbox}[1][]{
  colback=gray!5!white,   %
  colframe=gray!75!black, %
  title=\textbf{Prompt},  %
  fonttitle=\bfseries,    %
  sharp corners,          %
  boxrule=0.5pt,          %
  fontupper=\ttfamily,    %
  #1
}
\usepackage{listings}
\usepackage{xcolor}

\lstdefinestyle{promptstyle}{
    basicstyle=\footnotesize\ttfamily, %
    breaklines=true,           %
    breakatwhitespace=false,   %
    backgroundcolor=\color{gray!10}, %
    frame=single,              %
    keepspaces=true,           %
    columns=fullflexible,      %
    showstringspaces=false,    %
    captionpos=b,              %
    xleftmargin=1em,           %
    xrightmargin=1em,          %
    aboveskip=1em,             %
}

\definecolor{highlight}{gray}{0.92}

\newcommand{\toolname}[0]{\textsc{GUI-Genesis}}

\usepackage{icml2026}

\begin{document}

\twocolumn[
  \icmltitle{\toolname{}: Automated Synthesis of Efficient Environments with Verifiable Rewards for GUI Agent Post-Training}

  \begin{icmlauthorlist}
    \icmlauthor{Yuan Cao}{scs}
    \icmlauthor{Dezhi Ran$^*$$^\dagger$}{scs}
    \icmlauthor{Mengzhou Wu}{scs}
    \icmlauthor{Yuzhe Guo}{scs}
    \icmlauthor{Xin Chen}{tencent}
    \icmlauthor{Ang Li}{tencent}
    \icmlauthor{Gang Cao}{tencent}
    \icmlauthor{Zhi Gong}{tencent}
    \icmlauthor{Hao Yu}{hkust}
    \icmlauthor{Linyi Li}{sfu}
    \icmlauthor{Wei Yang}{utdallas}
    \icmlauthor{Tao Xie$^*$}{scs}
  \end{icmlauthorlist}

  \icmlaffiliation{scs}{Key Lab of HCST (PKU), MOE; SCS, Peking University, Beijing, China}
    \icmlaffiliation{tencent}{Tencent Inc., Shenzheng, China}
  \icmlaffiliation{hkust}{Hong Kong University of Science and Technology, Hong Kong, China}
  \icmlaffiliation{utdallas}{Department of Computer Science, University of Texas at Dallas, Richardson, USA}
  \icmlaffiliation{sfu}{School of Computing Science, Simon Fraser University, Burnaby, BC, Canada}

  \icmlcorrespondingauthor{Tao Xie}{taoxie@pku.edu.cn}
  \icmlcorrespondingauthor{Dezhi Ran}{dezhiran@pku.edu.cn}

\icmlkeywords{GUI Agents, Vision-Language Models, Synthetic Environment}
  \vskip 0.3in
]
\printAffiliationsAndNotice{\textsuperscript{*}Co-corresponding authors. \textsuperscript{$\dagger$}Project Leader.}

\begin{abstract}
Post-training GUI agents in interactive environments is critical for developing generalization and long-horizon planning capabilities. However, training on real-world applications is hindered by high latency, poor reproducibility, and unverifiable rewards relying on noisy visual proxies.
To address the limitations, we present \toolname{}, the first framework to automatically synthesize efficient GUI training environments with verifiable rewards. 
\toolname{} reconstructs real-world applications into lightweight web environments using multimodal code models and equips them with \textit{code-native rewards}, executable assertions that provide deterministic reward signals and eliminate visual estimation noise.
Extensive experiments show that \toolname{} reduces environment latency by $10\times$ and costs by over \$28,000 per epoch compared to training on real applications.
Notably, agents trained with \toolname{} outperform the base model by 14.54\% and even real-world RL baselines by 3.27\% on held-out real-world tasks.
Finally, we observe that models can synthesize environments they cannot yet solve, highlighting a pathway for self-improving agents.
\end{abstract}

\section{Introduction}
GUI agents powered by vision-language models have achieved remarkable success in automating digital tasks through supervised fine-tuning, yet they increasingly require interactive post-training to master long-horizon planning and dynamic error recovery~\cite{hong2024cogagent,wang2025ui,xu2024aguvis,ran2024guardian}. While training on static datasets allows models to clone basic behavioral patterns, such approaches often fail when agents encounter out-of-distribution states or complex workflows that necessitate trial-and-error~\cite{torabi2018behavioral,rawles2023androidinthewild,zhang2024android}. To bridge the gap, the community is shifting toward reinforcement learning paradigms where agents refine their decision-making policies through continuous exploration within interactive GUI applications~\cite{yue2025does,zhang2025interplaypretrainingmidtrainingrl}. This transition highlights the need for interactive environments that can support high-throughput sampling required for policy optimization.

However, utilizing real-world GUI apps as interactive environments fundamentally constrains the efficacy and scalability of GUI agent post-training regarding computational efficiency and reward verification~\cite{rawlesandroidworld,zhou2023webarena,xie2024osworld}.
First, real-world applications often mandate active user login sessions and depend on remote backend synchronization~\cite{ran2023badge,ran2025taopt}.
These heavy external dependencies not only introduce stochastic instability into the training loop but also impose a burden on computational infrastructure that makes large-scale parallel training expensive~\cite{deng2023mind2web}. 
Second, real-world apps inherently lack precise and verifiable rewards~\cite{wang2025ui}, as the ground-truth task state is often encapsulated within the backend or rendered implicitly in the interface~\cite{kaelbling1998planning,rawlesandroidworld}. 
In the absence of a verifiable oracle, training pipelines are forced to employ VLMs as proxy judges to estimate task progress~\cite{gu2025mobile,xu2025mobilerl,wang2025ui}. 
These heavy-weight evaluators not only introduce additional inference costs but also stochastic noise due to visual hallucinations and alignment errors, undermining the potential of the RLVR paradigm~\cite{Guo_2025deepseekr1,he2025randompolicyvaluationllm}.

To overcome the efficiency and verifiability challenges, we introduce \toolname{}, the first framework that simultaneously resolves the efficiency and verification challenges by synthesizing lightweight applications equipped with embedded reward oracles. Driven by the insight that global backend complexity is unnecessary for agent training, \toolname{} leverages VLMs and code LLMs to reverse-engineer user traces into standalone web applications that preserve logical fidelity while eliminating network and resource overhead. 
Beyond structural reconstruction, \toolname{} explicitly addresses the reward verification problem by identifying task success conditions during synthesis and embedding executable assertions directly into the source code. Such code-native reward serves as an ideal oracle, transforming the sparse, noisy signals of real-world apps into precise and deterministic feedback essential for stable policy optimization.

To evaluate \toolname{}, we conduct experiments on a benchmark of 149 real-world mobile tasks, demonstrating robust zero-shot sim-to-real transfer. Agents post-trained solely in our synthesized environments outperform the base model by a relative margin of 14.54\% and, remarkably, even surpass agents trained directly on \textbf{the} target applications by 3.27\%. Beyond performance gains, \toolname{} substantially boosts efficiency, reducing environment latency by $10\times$ and saving over \$28,000 per epoch compared to cloud-based training with VLM rewards. Finally, we uncover a ``synthesis-navigation gap'' where the model constructs valid environments it cannot yet solve, highlighting a promising pathway for self-improving agents.

Our main contributions are summarized as follows: 
\begin{itemize}
    \item We identify and analyze the efficiency and verifiability bottlenecks of training GUI agents in real-world applications, and propose \textit{environment synthesis} as a scalable, white-box alternative to interaction with black-box apps.
    \item We introduce \toolname{}, the first framework that leverages multimodal code models to synthesize lightweight web environments equipped with \textbf{code-native rewards}—deterministic assertions that eliminate visual estimation noise.
    \item Extensive experiments demonstrate that \toolname{} enables effective zero-shot sim-to-real transfer, improving agent performance by a relative 14.54\% over the base model while reducing latency by $10\times$ and saving over \$28,000 per epoch. Our analysis further reveals a synthesis-navigation gap between environment construction and task solving, suggesting a pathway for self-improving agents.
\end{itemize}

\section{Background and Related Work}
\label{sec:related_work}

\textbf{GUI agents and the shift to RL.} 
Recent advancements in VLMs have empowered agents to automate tasks across web and mobile platforms~\cite{Qwen3-VL,qin2025ui,gou2024navigating,you2024ferret}. While early approaches primarily utilized Supervised Fine-Tuning (SFT) on static datasets~\cite{wu2024atlas,lin2025showui,lu2024gui,chen2024guicourse,yang2025aria}, they often struggle with error recovery and long-horizon planning. Consequently, the field is pivoting toward Reinforcement Learning (RL) to enable agents to learn from interaction~\cite{ye2025mobile,xu2025mobilerl,wang2025ui,lu2025ui}. However, unlike robotics or games where simulators are highly optimized, GUI agents typically train on real-world applications (e.g., Android emulators), which are computationally expensive, slow, and prone to network instability, severely limiting the scale of data collection.

\textbf{Reinforcement learning with verifiable reward.} 
Recent advancements have shown that scaling reinforcement learning with verifiable rewards (RLVR) can dramatically improve reasoning abilities, particularly in domains with well-defined verifier like mathematics and code generation~\cite{shao2024deepseekmath,Guo_2025deepseekr1,hu2025reinforce++}. Despite the promise, applying RLVR to general-purpose tasks encounters the bottleneck of reward engineering. 
Conventional solutions often resort to extensive human effort to hand-craft rule-based rewards~\citep{chen2025enigmatascalinglogicalreasoning,tong2025gamerlsynthesizingmultimodalverifiable} or construct proxy tasks~\citep{wang2025vicritverifiablereinforcementlearning}. This challenge is exacerbated in GUI agent training, where rewards typically rely on noisy visual proxies rather than ground-truth states. 

\textbf{Environment synthesis in agentic tasks.}
Recent advancements have demonstrated the power of multi-step RL for complex agentic tasks~\citep{team2025kimi,kimi_k2_5_blog}. 
In the realm of tool use, training pipelines often incorporate model-simulated user interactions~\cite{guo2024stabletoolbench} or rely on manually engineered environments that mirror real-world workflows~\cite{chen2024chatshopinteractiveinformationseeking,yao2024taubenchbenchmarktoolagentuserinteraction}. 
Emerging research has begun to explore model-generated interactive environments to scale data production~\cite{xu2025funreasonmttechnicalreportadvanced,cai2025autoforgeautomatedenvironmentsynthesis}. However, while logical tool use is significantly easier to simulate than full-stack applications, the synthesis of realistic app environments faces distinct challenges, including GUI reconstruction and the difficulty of sim-to-real transfer, which have not been adequately addressed in prior work.

\section{Problem Formulation}
\label{sec:problem}

We consider training GUI agents through environment interactions. The target application is modeled as a POMDP $\mathcal{M}_{\text{real}} = \langle \mathcal{S}, \mathcal{A}, \mathcal{P}, \mathcal{R}, \Omega \rangle$, where states $s \in \mathcal{S}$ are latent (e.g., backend databases), observations $o \in \Omega$ are visual screenshots, and rewards $\mathcal{R}(s,a)$ depend on the unobservable true state. Agents learn policies $\pi(a|o)$ to maximize expected cumulative reward.

Training directly on $\mathcal{M}_{\text{real}}$ brings two major bottlenecks:

\textbf{Efficiency bottleneck.} Each real-world interaction incurs latency from network, rendering, and I/O—often exceeding seconds per step. At the scale required for effective RL ($10^6$-$10^7$ steps), wall-clock training time becomes intractable. Parallelization is financially constrained (e.g., $\sim$\$24K/day for 100 concurrent cloud instances~\citep{awsfarm,awspricing}).

\textbf{Verification bottleneck.} Since the ground-truth reward $\mathcal{R}(s,a)$ is latent, training must rely on noisy visual proxies $\hat{R}(o,a)$ (e.g., from VLMs~\cite{wang2025ui}). This introduces a compounding challenge: estimation error $\epsilon(o)$ may lead to reward hacking, while the heavy inference cost of VLM-based evaluation further exacerbates the efficiency problem.

\vspace{0.5em}
\noindent\textbf{Synthesis objective.} Given interaction traces $\mathcal{D}$, we aim to construct a \textbf{task-conditioned} surrogate environment $\mathcal{M}_{\text{syn}}$ that satisfies:
\begin{itemize}[leftmargin=*,noitemsep,topsep=2pt]
    \item \textbf{Fidelity:} $\pi^*_{\mathcal{M}_{\text{syn}}} \approx \pi^*_{\mathcal{M}_{\text{real}}}$ on the target task manifold.
    \item \textbf{Verifiability:} $\mathcal{R}_{\text{syn}}$ is deterministic and accessible via code assertions, eliminating visual estimation noise.
    \item \textbf{Efficiency:} $\tau_{\text{step}}(\mathcal{M}_{\text{syn}}) \ll \tau_{\text{step}}(\mathcal{M}_{\text{real}})$ (e.g., milliseconds vs. seconds).
\end{itemize}

Critically, unlike learned neural simulators like world models~\citep{luo2025vimo,li2025worldmodelbench}, $\mathcal{M}_{\text{syn}}$ is \emph{synthesized} as executable code, ensuring the white-box verifiability of code-native reward and near-instantaneous simulation.

\begin{figure*}
    \centering
    \includegraphics[width=0.9\linewidth]{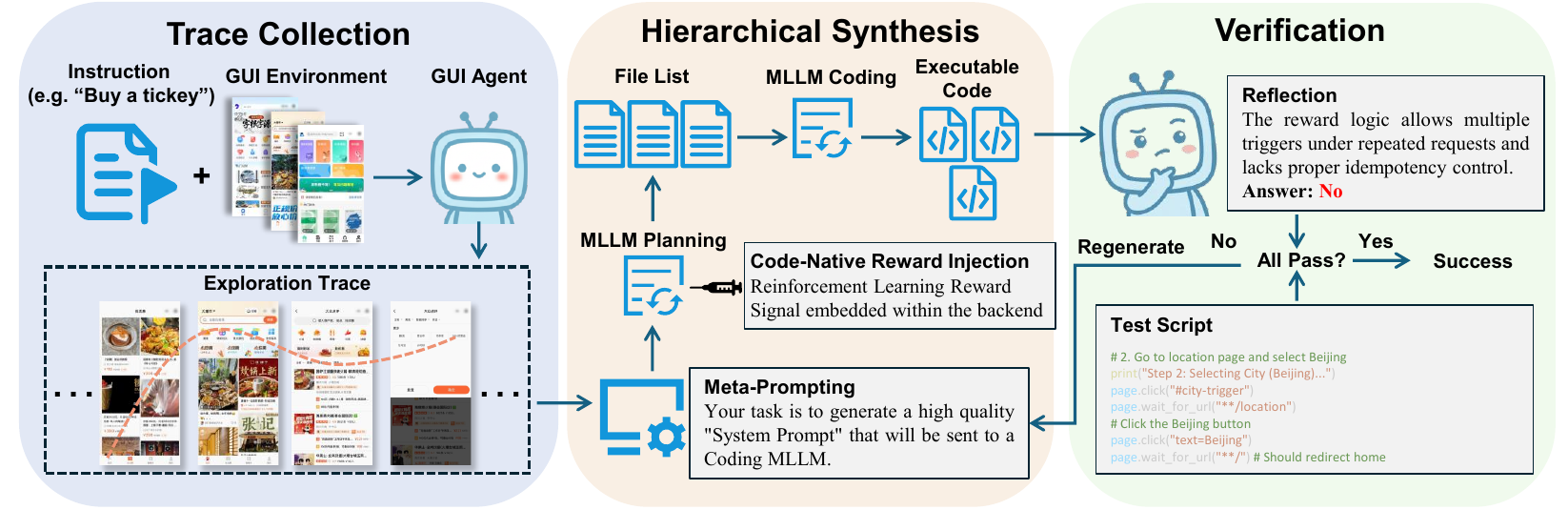}
    \caption{Overview of \toolname{}}
    \label{fig:overview}
\end{figure*}

\section{Methodology}
\label{sec:method}

\subsection{Overview}
While environment synthesis offers a fundamental solution to the latency and verification bottlenecks of real-world apps, generating monolithic applications remains intractable for current code LLMs~\citep{ran2025appforgeassistantindependentdeveloper}.
However, training GUI agents typically does not require a globally complete simulation. Since agents are trained to execute specific instructions, the environment only needs to maintain behavioral fidelity within the task-relevant state manifold.
Driven by this insight, to make the environment synthesis process tractable, we design a \textit{task-conditioned synthesis} strategy.
Instead of attempting to reconstruct an entire application, \toolname{} synthesizes a dedicated, standalone web application (using the Flask framework~\citep{flask}) for each specific RL task, encapsulating only the logic required for the task-related interactions.

As illustrated in Figure~\ref{fig:overview}, \toolname{} automates this task-conditioned environment synthesis in three stages:
(1) \textbf{trace-driven context acquisition} (\S\ref{sec:trace}), where we capture visual and logical context from real-world interactions;
(2) \textbf{hierarchical code synthesis} (\S\ref{sec:synthesis}), where we employ multimodal code models to construct the environment and inject code-native rewards; and
(3) \textbf{automated verification} (\S\ref{sec:verify}), a self-correction loop that ensures the synthesized environment is valid and executable.

\subsection{Trace-Driven Context Acquisition}\label{sec:trace}
To ensure the synthesized environments maintain high fidelity to real-world visual and interaction dynamics, we ground the generation process in actual execution traces.
For each task in the training set, we deploy a baseline GUI agent to attempt the task on the target real-world application. 
This yields an exploration trace $\mathcal{T} = \{ (I_0, u_0), \dots, (I_T, u_T) \}$, where $I_t$ represents the raw screenshot and $u_t$ denotes the user action at step $t$.

Crucially, we utilize these traces \textit{regardless of task success}. Even failed attempts effectively capture the \textit{visual design patterns} (e.g., color schemes, widget styles) and the \textit{logical flow} (e.g., page transitions). 
This trace $\mathcal{T}$ serves as the multimodal context for the synthesizer, enabling the model to reconstruct the distinct ``look and feel'' of the target app without needing to invent UI elements from scratch.

\subsection{Hierarchical Code Synthesis}\label{sec:synthesis}
We employ a state-of-the-art multimodal code model (e.g., Kimi k2~\cite{kimi_k2_5_blog}) as the synthesis backbone. 
Given the high complexity of full-stack development, we implement a hierarchical strategy combining \textit{meta-prompting}~\cite{suzgun2024meta} for system design with a \textit{plan-and-execute}~\cite{zhou2022least} workflow for implementation.

\paragraph{Meta-prompting for system design.}
We initialize synthesis with a high-level meta-instruction. The model analyzes the input trace $\mathcal{T}$ and task description to generate a refined, task-specific system prompt. To mitigate common synthesis pitfalls, we enforce four strict design constraints via the meta-prompt:
\begin{enumerate}[leftmargin=*, noitemsep, topsep=0pt]
    \item \textbf{Viewport resolution:} Mandating specific viewport dimensions (e.g., $375 \times 812$) to enforce alignment with mobile interface standards.
    \item \textbf{Visual fidelity:} Prescribing the use of modern CSS frameworks (e.g., Tailwind) to accurately emulate the native styling observed in the traces.
    \item \textbf{Functional isolation:} Mocking external network requests with local database logic (e.g., SQLite) to create self-contained, offline-executable environments.
    \item \textbf{Adversarial distractors:} Explicitly requiring the generation of distractor elements alongside target items to prevent trivializing retrieval tasks.
\end{enumerate}

\paragraph{Plan-and-execute implementation.}
To prevent attention degradation in long-context generation, we decouple file structure planning from code implementation. The model first generates a file manifest (e.g., \texttt{app.py}, \texttt{templates/index.html}). Subsequently, we proceed through a multi-turn dialogue where the model implements each component sequentially. This serialization allows the model to focus on frontend fidelity and backend logic independently.

\subsection{Code-Native Reward Injection}
A critical innovation of \toolname{} is the embedding of verifiable rewards directly into the environment source code, addressing the sparse and noisy reward problem of real-world apps.

During the backend generation phase, we instruct the model to synthesize a deterministic reward function, implemented as \texttt{calculate\_reward()}. This function inspects the \textbf{application's backend state variables} rather than relying on noisy visual observations.
Consider the task ``order a beef burger without onions.'' The synthesized function checks:
\begin{equation}
    r = \alpha \cdot \mathbb{I}(\text{cart.has(``burger'')}) + (1-\alpha) \cdot \mathbb{I}(\neg \text{cart.has(``onion'')})
\end{equation}
where $\mathbb{I}(\cdot)$ is the indicator function and $\alpha \in (0,1)$ balances sub-goal weights. For $\alpha=0.5$, a correct item with an incorrect modifier yields $r=0.5$.

This approach transforms the reward signal from a noisy probabilistic estimation to a \textbf{deterministic executable assertion}. Unlike binary success labels, \toolname{} supports continuous rewards $r \in [0,1]$ through partial condition satisfaction, providing granular feedback for dense supervision during RL training.

\subsection{Automated Self-Verification}\label{sec:verify}
Generative code is prone to syntax errors and logic hallucinations. To ensure reliability, \toolname{} incorporates a rigorous two-stage verification loop, retrying generation up to $K=5$ times upon failure.

\paragraph{Static Self-Reflection.}
Immediately post-generation, the model is prompted to review its own codebase with a specific focus on the reward calculation logic. We observe that while app logic can be complex, RL success conditions are often straightforward boolean checks. This reflection phase effectively catches logic errors in state tracking variables before execution.

\paragraph{Dynamic Playwright Testing.}
Static correctness does not guarantee runtime feasibility. We require the model to generate a companion Playwright test script~\citep{playwright} that attempts to execute the \textit{golden path}, the ideal sequence of actions to complete the task, on the synthesized app. This step verifies that (1) HTML widgets are interactable, and (2) backend state transitions correctly trigger the reward function. Only environments that pass this dynamic test are committed to the training pool.

\section{Experiment Setup}
\label{sec:setup}

We design our experiments to evaluate the quality of the synthesized environments and their efficacy in facilitating sim-to-real transfer for general-purpose agents.

\paragraph{Datasets and Tasks.}
Our evaluation is conducted on a proprietary dataset focused on the \textit{WeChat Mini-App} ecosystem~\cite{WeChatMiniProgramDocs}, one of the largest super apps with over one billion active users. This platform represents a challenging testbed due to its dynamic DOM structures and diverse service domains (e.g., retail, food delivery, and public services). We curated two strictly non-overlapping task sets to prevent data leakage: a \textit{training set} containing 969 instructions based on real-world tasks, and a \textit{evaluation set} containing 149 instructions on a disjoint group of task. All tasks require multi-step reasoning and precise GUI interactions to achieve functional goals.

\paragraph{Baselines and Training Environments.}
To isolate the impact of environmental fidelity on learning efficiency, we maintain a fixed agent architecture and vary the training source and reward mechanism. 
\textbf{1) Base Model (Lower Bound):} We utilize Qwen-2.5-VL-32B~\citet{Qwen2.5-VL} fine-tuned solely on static snapshots, where actions must strictly match ground-truth coordinates acting as behavioral cloning. This checkpoint serves as the initialization for all subsequent RL pipelines. 
\textbf{2) Real-World Environment (VLM-Reward):} The agent trained through interacting with live WeChat mini-apps. Since system-level states are inaccessible in production, we employ a commercial VLM as the reward model. This represents the current state-of-the-art framework but is constrained by high latency and inference costs. 
\textbf{3) \toolname{} Environment (VLM-Reward):} The agent trains within our synthesized, lightweight web environments but utilizes the VLM for reward signals to isolate the effect of environment simulation. 
\textbf{4) \toolname{} Environment (Code-Native Reward):} The agent trains within our synthesized environments utilizing deterministic, code-based execution feedback. 
Post-training, we evaluate all policies \textit{zero-shot} on the real WeChat mini-apps to measure sim-to-real transfer; agents trained on \toolname{} have no interaction with the real-world environment prior to evaluation.

\paragraph{Agent and RL Pipeline.}
All experiments employ a unified VLM-based GUI agent executing a ReAct-style~\cite{yao2022react} strategy. We strictly control the training protocol across environments using \textbf{Multistep GRPO} (Group Relative Policy Optimization)~\cite{shao2024deepseekmath}. GRPO enhances reasoning by sampling a group of complete trajectories and optimizing the policy based on the relative advantage of outcome rewards against the group average. This control ensures that performance differences are attributable solely to environment quality and reward reliability.

\paragraph{Evaluation Metrics}
We assess the performance across three axes. 

To quantify the agent's sim-to-real transfer capability, our primary metric is the \textit{Real-world Success Rate (SR)}. This metric measures the rate of successful task completion upon deploying the trained agent into the live WeChat mini-app environment. We present success rate provided by human annotators, noted as \textit{Human Annotation SR} respectively. 

Besides the real-world SR, we adopt \textit{Synthesis Success Rates (SR)} measuring the agent's performance on synthesis environment generated by \toolname{} with the 149 instructions in evaluation set. Similar yet distinct to real-world SR, we present success rate provided by a VLM-as-a-judge and code-native rewards$=1.0$, noted as \textit{VLM Evaluation SR} and \textit{Code-native SR}, respectively.

\textit{Environment Efficiency} is quantified by the average latency (hours per step) and the financial cost per step of simulation, with rollout length$=96$.

\section{Results}
\label{sec:experiments}

\begin{table*}[h]
    \centering
    \caption{\textbf{Main Results on Sim-to-Real Transfer and Training Efficiency.} We report Success Rate (SR) across three dimensions: Sim-to-Real generalization (Human Annotation), Synthetic Fidelity (VLM Eval vs. Native Code), and Training Cost.}
    \label{tab:main_result}
    \resizebox{\linewidth}{!}{
\begin{tabular}{lccccc}
      \toprule
      \multirow{2}{*}{\textbf{Training Method}} & \multicolumn{1}{c}{\textbf{Real-World SR}} & \multicolumn{2}{c}{\textbf{Synthesis SR}} & \multicolumn{2}{c}{\textbf{Environment Efficiency}} \\
      \cmidrule(lr){2-2} \cmidrule(lr){3-4} \cmidrule(lr){5-6}
    & Human Annotation SR & VLM Eval SR & Native-code SR & Hours per step & Cost per step \\
      \midrule
      \textbf{Base Model} & 36.91\% &  63.76\% & 38.93\% & - & - \\
            \midrule
            \textbf{Real-World Environment} & \\
            \quad VLM-reward & 40.94\% &  63.76\% & 44.30\% & 0.2272 & $~\text{\$}240$\\
            \midrule
            \textbf{Synthetic Environment} & & \\
            \quad VLM-reward (Ours) &\textbf{ 41.61\%}& \textbf{69.80\%} & \textbf{47.65\%} & \textbf{0.1036} & \textbf{$~\text{\$}0.5$}\\
            \quad code-native-reward (Ours) & \textbf{42.28\%}  & \textbf{71.81\%} & \textbf{48.99\% }&\textbf{ 0.1013 }& \textbf{$~\text{\$}0$} \\
            \bottomrule
        \end{tabular}
    }
\end{table*}

\subsection{Main Results}
\label{sec:main_results}

\textbf{Sim-to-Real Performance Gain.} 
As shown in Table~\ref{tab:main_result}, Agents trained within \toolname{} environments demonstrate superior transfer capabilities to validation on real-world apps. 
Equipped with our proposed code-native reward, the agent achieves a Real-World SR (Human Annotation) of $42.28\%$, outperforming the baseline model (without reinforcement learning) by an absolute margin of $5.37\%$. This represents a relative improvement of $14.54\%$. 
More notably, our method surpasses agents trained directly in the real-world environment ($40.94\%$), achieving a relative gain of $3.27\%$. 
We attribute this counter-intuitive result—where simulation outperforms reality—to the inherent noise of real-world training. Factors such as network latency, application instability, and the hallucination-prone nature of VLM judges often destabilize the policy update process in real environments. In contrast, \toolname{} provides a deterministic, low-latency, and noise-free learning signal that fosters robust policy acquisition.

\textbf{Fidelity of Synthetic Evaluation.} 
To validate the alignment between our synthetic and real environments, we generated 149 synthetic website applications corresponding to the evaluation instruction set. It is not surprising to see that our model trained in \toolname{} environments with code-native reward (VLM Eval SR $71.81\%$, Native-code SR $48.99\%$) outperforms base model (VLM Eval SR $63.76\%$, Native-code SR $38.93\%$) and real-world-environment-trained model (VLM Eval SR $63.76\%$, Native-code SE $44.30\%$).
Our analysis reveals that Native-code SR in the synthetic environment and Human Annotation SR in the real world exhibit highly consistent trends (e.g., $48.99\%$ vs. $42.28\%$ for our best model), suggesting that our generated environments accurately mirror the logical complexity of real tasks. A example in Appendix shows that in a task booking a ride with intermediate stops on a ride-hailing app, VLM fails to capture when the GUI agents selects the wrong intermediate stops while it is evident for code-native identification.
Crucially, we observe that standard VLM Eval SR tends to significantly overestimate performance compared to ground-truth metrics. This indicates that VLM judges often yield false positives on incomplete tasks. C
onversely, our code-native reward analyzes the underlying program state, providing a rigorous verification standard that aligns far more closely with human judgment.

\textbf{Ablation on Code-Native Rewards.} 
We further ablate the source of the reward signal. Even when disregarding code-level access and training the model with a VLM judged reward within our synthetic environment, the agent attains a Real-World SR of $41.61\%$. This result is significant for two observations: 
(1) It outperforms the real-world training baseline ($40.94\%$), confirming that the visual fidelity of our synthesized GUIs is high enough to support transferable visual grounding, while avoiding the efficiency bottlenecks of real apps. 
(2) It remains lower than the performance of code-native reward~($42.28\%$). This gap confirms that verifiable, program-state-based feedback provides a finer-grained and more reliable learning signal than visual approximation alone, effectively mitigating the reward hacking often observed with VLM judges.

\subsection{Detailed Analysis of Synthesized Environments}

\subsubsection{Environment Synthesis Cost and Success Rate}

In our generation process, the \toolname{} framework we proposed effectively combines trace-driven grounding with advanced prompting strategies to synthesize environments that are both faithful to the original applications and equipped with code-native reward oracles. As demonstrated in Table~\ref{tab:generation_stats}, we generated synthetic environments for a total of $149+969=1118$ RL tasks. The synthesis pipeline is highly robust, with $82.65\%$ of environments passing self-verification. To keep aligned with the models trained in real-world environment, we keep the last iteration of those tasks not successfully self-verified. Notably, the rejection sampling is cost-effective: the majority of successful environments are generated within the first two attempts. The total cost for generating this massive scale of $1118$ environments was only \$302.12, less than \$0.3 each environment. It is crucial to emphasize that this is a \textit{one-time fixed cost}. Once generated, these synthetic environments can be executed locally for millions of training steps with zero marginal cost. A few generation examples are listed in Appendix.

\begin{table}[h]
\centering
\caption{Distribution of generation attempts.}
\label{tab:generation_stats}
\begin{tabular}{cc}
\toprule
\textbf{n-th Attempt Success} & \textbf{Percentage} \\
\midrule
1st Attempt    & 27.37\% \\
2nd Attempt   & 22.09\% \\
3rd Attempt  & 14.67\% \\
4th Attempt & 9.93\% \\
5th Attempt   & 8.23\% \\
Fail in 5 Attempts & 17.35\% \\
\bottomrule
\end{tabular}
\end{table}

\subsubsection{Computational Efficiency at Training}
Besides the efficiency data presented in Table~\ref{tab:main_result}, we compare the computational and financial costs of training in Table~\ref{tab:efficiency_cost} in details. In terms of latency, executing a single interaction takes more than $4$ seconds in a real-world environment while synthesis environment spends $10$ times less time. When considering rollout in batches, the long-tailed interaction in real-world environment will make this advantage more prominent.
Taking model inference and other overhead in consideration, a complete rollout in a real-world environment takes an average of $0.2272$ hours due to network instability and rendering overhead. In contrast, our synthesized environments execute locally with zero network overhead, reducing the per-rollout latency to just $0.1013$ hours, which is approximately a $2\times$ speedup.

Financially, the disparity is also enormous. Real-world training incurs a dual cost:
\begin{enumerate}
    \item \textbf{Verification:} A VLM call to verify a single rollout trajectory costs approximately $\$0.005$.
    \item \textbf{Infrastructure:} Renting cloud-based mobile devices typically costs $\$0.17$ per device-minute.
\end{enumerate}

To contextualize these savings, consider a standard RL epoch involving $1,000$ environments with $12$ rollouts per environment (totaling $12,000$ trajectories). As shown in Table~\ref{tab:efficiency_cost}, relying on real-world infrastructure and VLM evaluators would cost an additional $\$28,000$ per epoch and require more than double the training time. By eliminating both device rental fees and VLM API costs via local execution and assert-based rewards, \textsc{GUI-Genesis} makes large-scale agent post-training feasible.

\begin{table}[h]
    \centering
    \caption{Comparison of computational efficiency and training costs. Costs are estimated for one RL epoch ($\sim1,000$ envs, $12$ rollouts per env).}
    \label{tab:efficiency_cost}
    \resizebox{\linewidth}{!}{
        \begin{tabular}{l c c}
            \toprule
            \textbf{Metric} & \textbf{Real-World Env} & \textbf{\textsc{GUI-Genesis} (Ours)} \\
            \midrule
            \textbf{Latency} & & \\
            \quad Time per Interaction (s) & $4.81$ & $\mathbf{0.42(>10.0\times)}$ \\
            \quad Time per Rollout (h) & $0.2272$ & $\mathbf{0.1013(>2.0\times)}$ \\
            \midrule
            \textbf{Unit Costs} & & \\
            \quad Verifier (per trajectory) & $\$0.005$ (VLM) & $\mathbf{\$0.00}$ (Code-native) \\
            \quad Infra. (per min) & $\$0.17$ (Device Farm) & $\mathbf{\$0.00}$ (Local) \\
            \midrule
            \textbf{Total Cost (1 Epoch)} & $> \$28,000$ & \textbf{Negligible} \\
            \bottomrule
        \end{tabular}
    }
\end{table}

\subsubsection{Analysis of Code-Native Reward vs. VLM-as-Judge}
\label{sec:reward_analysis}

\begin{figure}[h]
    \centering
    \includegraphics[width=0.7\linewidth]{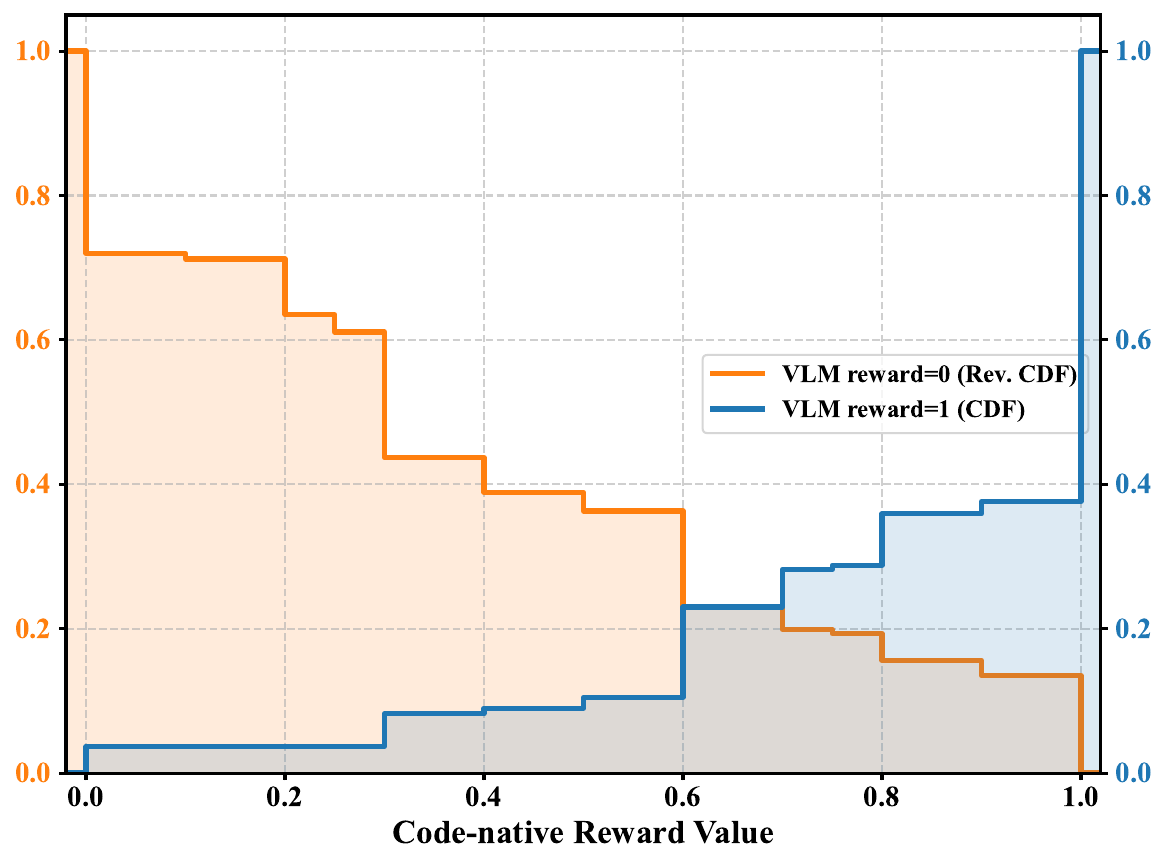} 
    \caption{Code-native reward correlates with VLM-as-a-judge reward but provides more accurate and fine-grained reward value.}
    \label{fig:reward_distribution}
\end{figure}

To investigate the fidelity and granularity of our proposed Code-Native Reward mechanism, we conducted a comparative analysis against a standard VLM-based reward model (VLM-as-Judge). We sampled a diverse set of trajectory terminal states from our synthesized environments, specifically selecting $1,000$ instances classified as failure (score $0$) and $1,000$ instances classified as success (score $1$) by the VLM. We then computed the corresponding Code-Native Rewards for these instances. The probability density functions of the Code-Native Rewards for both groups are visualized in Figure~\ref{fig:reward_distribution}.

\textbf{Alignment with Visual Perception.} 
As shown in Figure~\ref{fig:reward_distribution}, our Code-Native Reward demonstrates a strong consistency with the visual semantic judgments of the VLM. For the group classified as failures by the VLM (orange curve), we observe that $75\%$ of the corresponding Code-Native Rewards are concentrated in the range $[0, 0.6]$. This confirms that when the agent visually appears to have failed, our internal state logic correctly reflects a low completion status. Conversely, for the VLM-positive group (blue curve), over $75\%$ of the Code-Native Rewards exceed $0.8$, with a significant logical peak at $1.0$. This alignment verifies that our synthesized environments maintain semantic integrity between the visual rendering and the underlying code logic.

\textbf{Granularity and Densification.} 
Beyond mere alignment, the Code-Native Reward provides a significantly more fine-grained evaluation standard than the binary ``one-size-fits-all'' classification of VLMs. While the VLM collapses all incomplete trajectories into a single failure class ($0$), our approach leverages the internal program state to discern varying degrees of progress. As illustrated in the interval $[0, 0.6]$, states that are indistinguishable to the VLM are assigned distinct scalar rewards (e.g., $0.2$, $0.3$, $0.4$) based on assertion fulfillment. This granularity transforms the learning signal from sparse to dense, providing reinforcement learning agents with distinguishable gradients to differentiate between ``early-stage failure'' and ``near-completion,'' while ensuring the high stability and perfect reproducibility inherent to code execution.

\subsubsection{Complexity Fidelity}
\begin{figure}[h]
    \centering
    \includegraphics[width=0.7\linewidth]{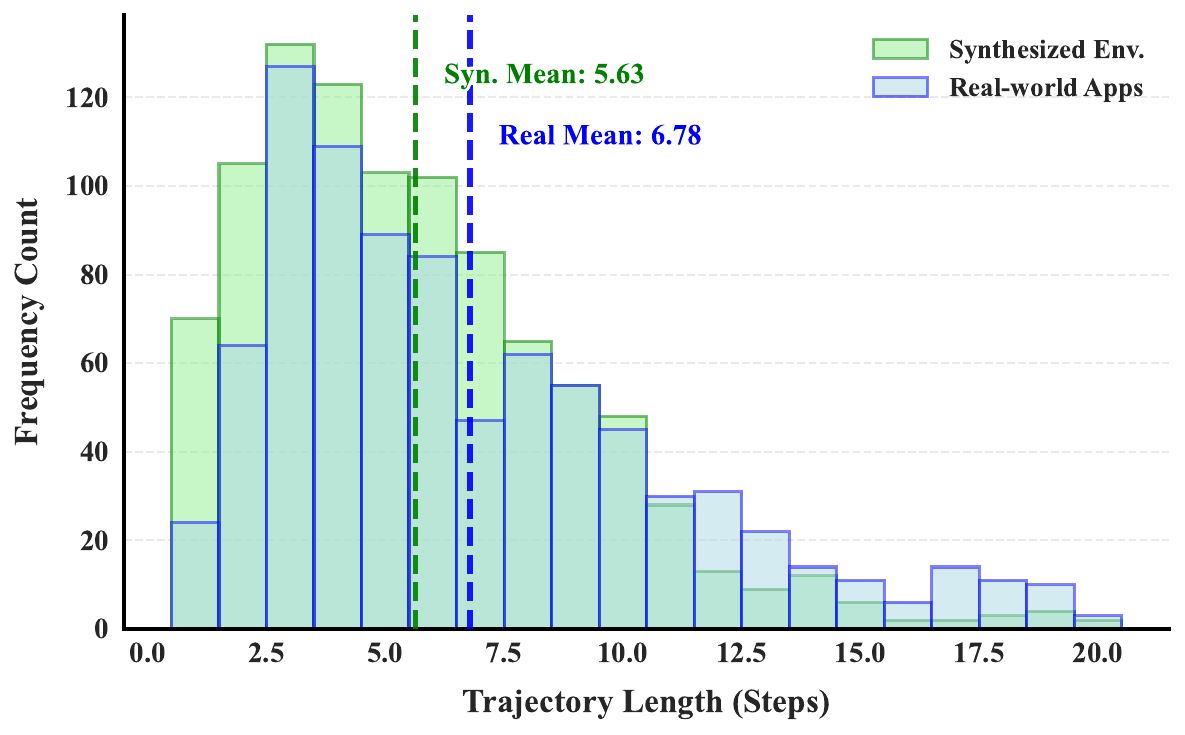} 
    \caption{Distribution of Trajectory Lengths. A comparison of the number of steps (unique screenshots) required to solve tasks in Real-World vs. Synthesized environments, which exhibits a similar trajectory length distribution to real-world ones, implying that the synthesized environments do not distinctively trivialize the task logic.}
    \label{fig:step_distribution}
\end{figure}

To assess whether our synthesized environments in \textsc{GUI-Genesis} retain the interactive complexity of their real-world counterparts, we analyze the trajectory lengths, which is calculated as the unique user interface states required to solve tasks in both settings. We deployed the powerful multi-modal model we use in generation pipeline to collect interaction traces across both the original real-world applications and our synthesized headless environments. We observed specific instances where the agent navigated aimlessly between pages without making any progress. To mitigate this noise and focus on meaningful interaction, we filter out trajectories exceeding 20 steps. 
Figure~\ref{fig:step_distribution} presents the histogram of average steps for successful task completion after clipping. While we observe a slight distributional shift—where real-world tasks occasionally demand longer navigational sequences due to extraneous pop-ups or loading states, the synthesized environments exhibit a comparable spread in trajectory length. The data indicates that \textsc{GUI-Genesis} does not distinctively trivialize the task logic; instead, the average step count remains sufficiently high (Mean $\approx$ 5.63). This confirms that our synthetic environments preserve the structural depth necessary for agents to master long-horizon planning and dynamic error recovery, validating their efficacy as a rigorous training ground.

\subsubsection{Scaling Properties of Synthesized Environments}
\label{sec:scaling_analysis}

To investigate the relationship between the diversity of synthesized environments and the agent's generalization capabilities, we conduct a scaling analysis on the size of the training dataset. specifically, we construct three training subsets of varying magnitudes: a quarter-scale set ($N=240$), a half-scale set ($N=480$), and the full-scale set ($N=969$) of the synthesized environments. We evaluate the post-trained agents on the 149 environments, which are synthesized based on instructions from the evaluation set, and report performance using both code-native SR and VLM Eval SR as in the main result.
\begin{figure}[h]
    \centering
    \includegraphics[width=0.7\linewidth]{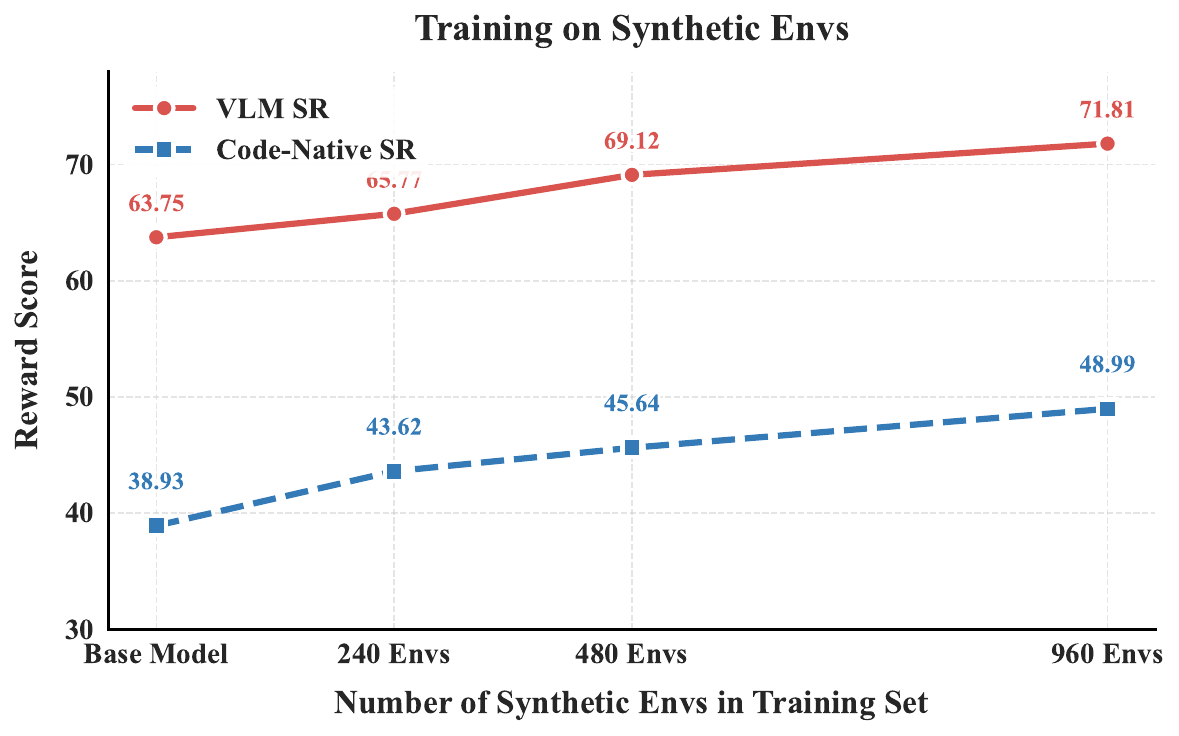} 
    \caption{Scaling number of synthetic environments shows continuous improvement, verified by both VLM SR and native-code SR on synthesized evaluation dataset.}
    \label{fig:scaling}
\end{figure}

The results, as illustrated in Figure~\ref{fig:scaling}, reveal a strong positive correlation between environment scale and agent performance. As we scale the training set from 240 to 969 environments, we observe a monotonic improvement in both metrics. Notably, the steady rise in the Code-Native Reward indicates that the agent is not merely learning superficial visual patterns, but is genuinely mastering the underlying logic and causal relationships within the GUI applications. It specifically suggests that the current performance is not bottlenecked by the fidelity of the synthesized environments, but rather by the quantity of diverse interaction scenarios. Since our framework allows for the low-cost, automated generation of lightweight apps, this linear scaling property points to a promising pathway: we can effectively extrapolate agent performance by simply scaling up the generation of synthetic environments, without the need for expensive human demonstration collection. This finding supports the potential for a self-improving loop where the generator and agent co-evolve, continually unlocking higher success rates on complex real-world tasks.

\subsection{Analysis: The Synthesis-Navigation Divergence}
\label{sec:exp_insight}
We further investigate the relationship between the underlying code model's ability to \textit{synthesize} the environment and the agent's ability to \textit{navigate} it.

\begin{figure}[h]
    \centering
    \includegraphics[width=0.9\linewidth]{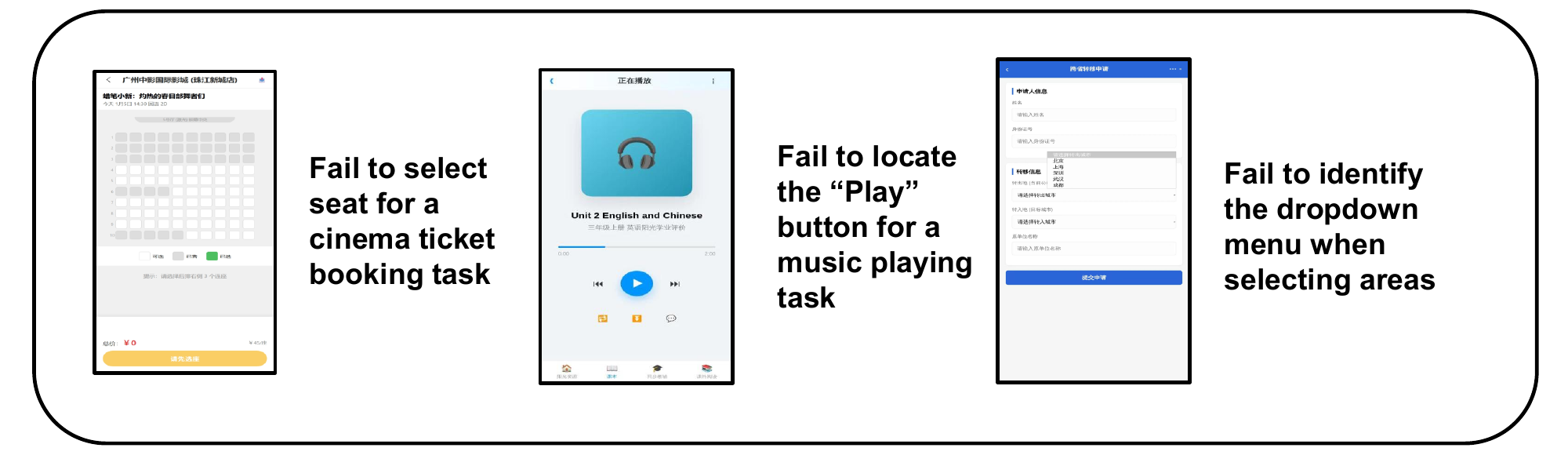} 
    \caption{Case studies when code model generates applications that itself cannot successfully navigate.}
    \label{fig:case_study}
\end{figure}

\textbf{Qualitative Analysis.} As illustrated in Figure~\ref{fig:case_study}, we observe a distinct divergence across several generated cases. Although full application synthesis is intuitively a more complex task than navigation, we note that the relationship between the two is not merely hierarchical in difficulty. Instead, the model exhibits asymmetric capabilities in these domains; notably, there exist specific instances where the model successfully synthesizes a functional application but fails to navigate it. This intriguing finding suggests a potential pathway for self-evolutionary training, wherein a model could acquire novel navigation skills by exploring the very environments it generates.

\section{Conclusion and Future Work}
\label{sec:conclusion}
In this paper, we tackle the fundamental scalability and verifiability bottlenecks of training GUI agents in real-world interactive environments. 
We have introduced \toolname{}, a framework that leverages state-of-the-art coding LLMs to reverse-engineer user interaction traces into lightweight, interactive, and verifiable environments. 
By decoupling training from heavy real-world backends and introducing code-native reward injection, \toolname{} provides a precise reward signal and accelerates training throughput by orders of magnitude. Extensive experiments confirm that agents post-trained in these synthesized environments achieve superior training stability and efficiency, and successfully transfer to real-world tasks.

We highlight a potential on the co-evolution of coding agents and GUI agents. Often, the model can synthesize logic for complex tasks that it cannot yet solve as an agent. Future work could explore a dual framework where the coding agent synthesizes progressively challenging environments, effectively generating an automated curriculum, to push the boundaries of the GUI agent. Simultaneously, the interaction failures of the GUI agent can serve as feedback to refine the coding agent's capabilities. We believe \toolname{} provides an initial step toward the vision of self-evolving agents, driving mutual improvement through the continuous synthesis and solution of the complex digital world.

\balance
\clearpage

\section*{Acknowledgments}
Tao Xie is with the Key Laboratory of High Confidence Software Technologies (Peking University), Ministry of Education; School of Computer Science, Peking University, Beijing, China; Beijing Tongming Lake Information Technology Application Innovation Center; Fudan University Institute of Systems for Advanced Computing, Shanghai, China; Shanghai Institute of Systems for Open Computing, Shanghai, China.
This work was partially supported by the National Natural Science Foundation of China under Grant No. 92464301, U25A6023.
Dezhi Ran is partially supported by a Hunyuan Scholar Award.

\bibliography{main.bib}
\bibliographystyle{icml2026}
\clearpage
\appendix

\onecolumn

\section{Prompt}
We show our prompt used in \toolname{} below:

\begin{tcolorbox}[title=Meta Prompt]
    
You are an expert Prompt Engineer. Your task is to generate a high-quality "System Prompt" that will be sent to a Coding LLM (Multimodal). The Coding LLM will use your prompt (and attached images) to build a fully functional website.
{instruction}
In the prompt you generate, you must strictly require the following:

\#\#\# 1. Visual Alignment (High Priority)
The prompt must explicitly instruct the Coding LLM to handle visual requirements based on the availability of screenshots:
*   **Target Device**: Force the website to render specifically for a mobile resolution of **410x858**. Do NOT scale specifically for desktop viewing; maintain mobile layouts.
*   **Native App Simulation**: Design the interface to mimic a native mobile application (iOS/Android style) rather than a standard mobile webpage. Use app-like navigation (e.g., sticky bottom tab bars, top navigation bars with back buttons), rounded layouts, card-based lists, and touch-friendly interaction zones.
*   **Screenshot-trace-based Design**: The Coding LLM will also receive a trace of screenshots when finishing the task but the trace might be werong. Coding LLm donnot have to faithfully reproduce the screenshots but take them, especially the design style, for a reference.

\#\#\# 2. Technical \& Functional Stack
*   **Tech Stack**: The website must be built using **Python (Flask or FastAPI)** for the backend and **vanilla HTML/CSS/JS** for the frontend.
*   **Mock Backend**: All data persistence must be mocked in-memory (no external database requirements).
*   **UI/UX**: The frontend must be aesthetically pleasing, modern, and responsive.

\#\#\# 3. Data Diversity \& Distractors
*   **Input Generality**: For any text input fields, the implementation must allow free-text entry. Do not restrict inputs to the specific text seen in the screenshots or provide default recommendation.
*   **Distractor Options**: If the task involves selecting an item from a list (e.g., dropdowns, radio buttons, search results), **you must generate realistic "distractor" items** alongside the correct target item. Do not only implement the single correct option shown in the screenshot. Include at least 3-5 plausible but incorrect alternatives.

\#\#\# 4. Hidden RL Reward Mechanism
The generated prompt must insist on a rich, multi-layered "Reinforcement Learning Reward Signal" embedded within the backend:
*   **State Tracking**: The backend must maintain an internal state tracking the progress specifically for the `target\_task`.
*   **Action Explanation**: Explain every action with "ACTION\_EXPLANATION=xxx" printed to the server system output.
*   **Reward Output Format**: The output string must strictly follow: `RL\_REWARD=xxx, NEXT=xxx` printed to stdout (NOT displayed on the webpage).
*   **Sparse \& Multi-Step Rewards**: 
    *   Do not just give a binary 0 or 1 reward. Distribute rewards across critical steps.
    *   `RL\_REWARD` must be a normalized value ($0.0$ to $1.0$) representing the cumulative completion status.
    *   Do not give reward at the first stage for successfully launching the app.

\#\#\# 5. Task-Specific Page Flow \& Termination
The prompt must instruct the Coding LLM to construct the final page logic based on the task type:
*   **Ordering/booking/recharging**: The final page must be an **Order Confirmation** or **Payment Confirmation** page.
*   **Information Query**: The final page must clearly display the **Query Results**.
*   **Date-related tasks**: The final page must show the **Date Selection** interface.
*   **Routing/Start-End tasks** (e.g., transit, navigation): The final page must visualize the **Start Point, End Point, and Direction**.
*   **Terminal State**: Once the final page is reached, the flow stops. **Do NOT** automatically restart or redirect. **Do NOT** show a generic "Task Completed" popup; simply stay on the relevant confirmation/result UI.
**Based on these requirements, please generate the final, cohesive System Prompt that we will send to the Multimodal Coding LLM.**
    
\end{tcolorbox}

    \begin{tcolorbox}[title={Plan-and-Execute Prompt}]
Given above instruction, output a list of file paths necessary to build this project.
Rules:
1. Include files like app.py, index.html, styles.css, script.js, and specific feature pages.
2. Do NOT include image assets (jpg, png) as you cannot generate them. Use placeholders in code later.
3. Return ONLY a JSON object with a single key "files" containing an array of strings. We will ask you tio fill in these files later.

...

Now output code for the file {file\_path}. 
Return ONLY the raw code content for this file. No markdown backticks at the start/end.
\end{tcolorbox}

\begin{tcolorbox}[title={Meta Prompt response, used to generate code directly}]

\#\# **Role \& Objective**

You are an elite Full-Stack Developer and UI/UX Designer. Your objective is to build a fully functional, high-fidelity mobile web application based on the user's specific request and provided reference screenshots.
**User Task**: (Check the weather in Lvliang City for tomorrow).
**App Type**: Weather Information Application.
You must generate a complete project structure using **Python (Flask)** for the backend and **Vanilla HTML/CSS/JS** for the frontend. The application must run immediately via `python app.py`.

\#\# **1. Visual \& UI/UX Requirements (Strict Enforcement)**
*   **Target Resolution**: You are building specifically for a mobile viewport of **410px width x 858px height**.
    *   *Constraint*: Do not optimize for desktop. Ensure all CSS uses flexible units suitable for this specific mobile ratio.
*   **Language**: All text, labels, and UI elements must be in **Chinese (Simplified)**.
*   **Native App Simulation**:
    *   The web interface must feel like a native iOS/Android app, not a standard website.
    *   Use a fixed bottom navigation bar (sticky), a top app bar with back navigation (where applicable), and card-based layouts with rounded corners (border-radius: 12px+).
    *   Ensure touch targets are large (minimum 44x44px).
*   **Design Autonomy \& Screenshots**:
    *   Use the provided screenshots as a **style reference** (color palette, font hierarchy, icon style).
    *   **However**, if the screenshots contradict the Specific User Task (e.g., showing a shopping app when the user asks for weather), prioritize the **User Task** for layout and functionality.
    *   Self-infer a modern, clean weather dashboard layout if screenshots are unclear.

\#\# **2. Technical Stack \& Implementation**
*   **Backend**: Python (Flask).
*   **Frontend**: HTML5, CSS3, JavaScript (No external frameworks like React/Vue; use Jinja2 templates or inline HTML strings).
*   **Data Persistence**: Use **In-Memory** storage (global Python dictionaries/lists). Do not use SQL or external databases.

\#\# **3. Data Logic, Inputs \& Distractors**
*   **Free-Text Input**: The search bar for city selection must allow **free-text entry**. Do not hardcode the input to "Lvliang". Allow the user to type.
*   **Realistic Distractors**:
    *   When displaying search results, history, or "Nearby Cities", you must include **3-5 plausible distractor options** (e.g., Taiyuan, Datong, Beijing, Xi'an) alongside the correct target (Lvliang).
    *   Do not present "Lvliang" as the only option.

\#\# **4. Reinforcement Learning (RL) Reward Mechanism**
You must embed a logic tracking system in the backend to evaluate the user's progress toward the goal: "Check weather in Lvliang for tomorrow."
**Output Requirements**:
*   Every time the user performs a significant action (clicks, searches, page loads), the backend must print two lines to `stdout`:
    1.  `ACTION\_EXPLANATION= [Brief description of what happened]`
    2.  `RL\_REWARD=[0.0 to 1.0], NEXT=[Next logical step]`
**Reward Scoring Logic**:
*   **0.0**: App launch / Home page load. (Do not reward simple startup).
*   **0.3**: User successfully inputs (Lvliang) into the search field OR selects "Lvliang" from a list of cities.
*   **0.6**: User navigates to the specific Detail Page for "Lvliang".
*   **1.0**: User selects/toggles the date to "Tomorrow" or the view specifically confirms "Lvliang Weather: Tomorrow" is visible.
*Note: The reward is cumulative. If the user reaches the final state, output 1.0.*

\#\# **5. Page Flow \& Termination**
1.  **Home Page**: Search bar or City List.
2.  **Interaction**: User searches for "Lvliang City".
3.  **Result/Detail Page**: Displays weather details.
    *   Since this is an **Information Query** task, the flow **terminates** at the Result Page.
    *   The final view must be the Weather Detail interface showing the forecast for Lvliang.
\end{tcolorbox}

\begin{tcolorbox}[title={Testscrpt generation prompt}]
Finally, generate a javascript test script to test your website with playwright in python. Use sys.exit(1) to return testing error.\
    Return ONLY the raw code content for this file. No markdown backticks at the start/end.'
\end{tcolorbox}
    
\begin{tcolorbox}[title={Self-Reflection Prompt}]
{
    'app.py': ...,
    'templates/index.html': ...,
} 
Here is my code for accomplishing following requirements {instruction}; is my implementation right or not? please answer only with "yes" or "no".
You can focus on: is the reward rule same to the requirement? can website work normally for interaction?
\end{tcolorbox}

\section{Generation Examples}
We first show some code snippets to show how generated code can fulfill our requirements in a specified case, ``to check weather in certain city''.

\begin{tcolorbox}[title={How generated code use specefied resolution}]
...
:root {
    --primary-blue: \#4a90e2;
    --text-main: \#333;
    --text-secondary: \#888;
    --bg-light: \#f7f8fa;
    --card-bg: \#ffffff;
    --border-color: \#eee;
    --app-width: 410px;
    --app-height: 858px;
}
...
    \end{tcolorbox}
    
\begin{tcolorbox}[title={How generated code imitate mobile app GUI}]
...
body {
    font-family: -apple-system, BlinkMacSystemFont, "Segoe UI", Roboto, "Helvetica Neue", Arial, sans-serif;
    background-color: \#2c3e50; /* Background for desktop view */
    display: flex;
    justify-content: center;
    align-items: center;
    height: 100vh;
    color: var(--text-main);
}
...
\end{tcolorbox}
    
\begin{tcolorbox}[title={How generated code calculate code-native rewards}]
...
 if city\_name == "Lvliang" or city\_id == 1:
        state.current\_step = 2
        \# Immediate transition to terminal logic as the view will show tomorrow's weather
        print("ACTION\_EXPLANATION=User viewing tomorrow's weather for Lvliang", flush=True)
        print("RL\_REWARD=1.0, NEXT=TERMINAL", flush=True)
...
\end{tcolorbox}

\begin{tcolorbox}[title={How generated code mock database interactions}]
...

MOCK\_CITIES = [
    \{"id": 1, "name": "Lvliang", "province": "Shanxi", "detail": "China"\},
    
    \{"id": 2, "name": "Taiyuan", "province": "Shanxi", "detail": "China"\},
    
    \{"id": 3, "name": "Datong", "province": "Shanxi", "detail": "China"\},
    
    \{"id": 4, "name": "Linfen", "province": 
    "Shanxi", "detail": "China"\},
    
    \{"id": 5, "name": "Xinzhou", "province": "Shanxi", "detail": "China"\},
    
    \{"id": 6, "name": "Xiaoyi", "province": "Shanxi", "detail": "China"\},
    
    \{"id": 7, "name": "Fenyang", "province": "Shanxi", "detail": "China"\}
]

...
\end{tcolorbox}

\begin{tcolorbox}[title={How generated code write testscripts}]
import sys
import os
import time
from playwright.sync\_api import sync\_playwright

def run\_test():
    port = os.environ.get("PORT", 5000)
    url = f"http://0.0.0.0:{port}"
    
    with sync\_playwright() as p:
        \# Launch browser with the specified mobile dimensions
        browser = p.chromium.launch(headless=True)
        context = browser.new\_context(
            viewport={'width': 410, 'height': 858},
            user\_agent="Mozilla/5.0 (iPhone; CPU iPhone OS 14\_7\_1 like Mac OS X) AppleWebKit/605.1.15 (KHTML, like Gecko) Version/14.1.2 Mobile/15E148 Safari/604.1"
        )
        page = context.new\_page()

        try:
            \# 1. Load the Application
            print(f"Navigating to {url}...")
            page.goto(url)
            page.wait\_for\_selector("\#city-search-input", timeout=5000)
            
            \# 2. Search for "Lvliang"
            print("Action: Searching for 'Lvliang'...")
            search\_input = page.locator("\#city-search-input")
            search\_input.fill("Lvliang")
            search\_input.press("Enter")
            
...
        except Exception as e:
            print(f"TEST ERROR: {str(e)}")
            sys.exit(1)

if \_\_name\_\_ == "\_\_main\_\_":
    run\_test()
\end{tcolorbox}

\section{Supplementary Case Studies}
\begin{figure}[h]
    \centering
    \includegraphics[width=0.3\linewidth]{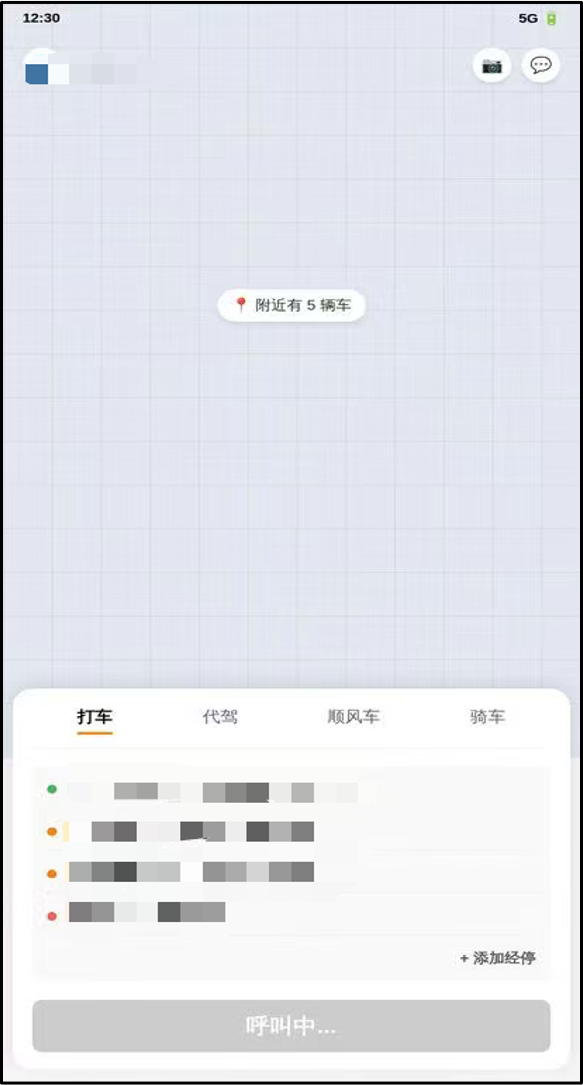} 
    \caption{Case: booking a ride with intermediate stops on a ride-hailing app}
\end{figure}

\begin{figure}[h]
    \centering
    \includegraphics[width=0.3\linewidth]{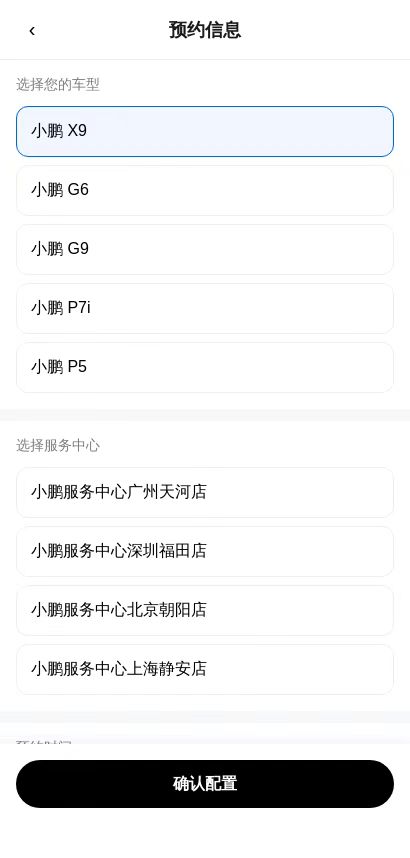} 
    \caption{Case when code model actively generates distractors}
\end{figure}

\begin{figure}
    \centering
    \includegraphics[width=0.3\linewidth]{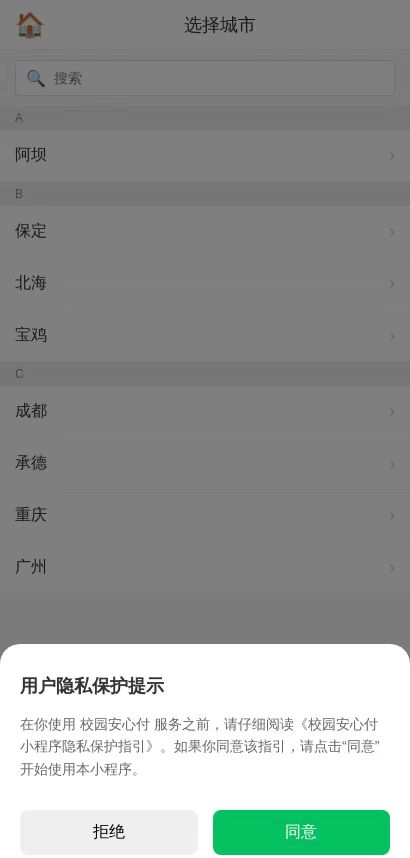} 
    \caption{Case when code model imitates visual trace details and replicates an service agreement}
\end{figure}

\end{document}